\documentclass[runningheads]{article}

\usepackage{arxiv}

\usepackage[utf8]{inputenc} 
\usepackage[ruled,vlined,linesnumbered,algoruled]{algorithm2e}
\usepackage{algorithmic}
\usepackage[T1]{fontenc}    
\usepackage{hyperref}       
\usepackage{url}            
\usepackage{booktabs}       
\usepackage{amsfonts}       
\usepackage{nicefrac}       
\usepackage{microtype}      
\usepackage{lipsum}
\usepackage{graphicx}
\usepackage[utf8]{inputenc}
\usepackage{caption}
\usepackage{float}
\usepackage{refstyle}
\usepackage[export]{adjustbox}
\usepackage{amsmath}
\usepackage{ulem}
\usepackage{fix-cm}
\usepackage{array}
\usepackage{caption}
\usepackage{subcaption}
\usepackage{longtable}
\usepackage{changepage}
\usepackage{textcomp}
\usepackage{multirow}
\usepackage{fancyhdr}
\usepackage{comment}

\newcommand\teamname{cs60075\_team2}

\begin{document}

\title{\teamname~at SemEval-2021 Task 1 : Lexical Complexity Prediction using Transformer-based Language Models pre-trained on various text corpora}



\author{Abhilash Nandy \qquad Sayantan Adak \qquad Tanurima Halder \qquad Sai Mahesh Pokala\\\texttt{\{nandyabhilash,sayantanadak.skni,haldertanurima,pokalasaimahesh\}@gmail.com}\\Indian Institute of Technology, Kharagpur, India}

\maketitle
\begin{abstract}
This paper describes the performance of the team \teamname~at SemEval 2021 Task 1 - Lexical Complexity Prediction. The main contribution of this paper is to fine-tune transformer-based language models pre-trained on several text corpora, some being general (E.g., Wikipedia, BooksCorpus), some being the corpora from which the CompLex Dataset was extracted, and others being from other specific domains such as Finance, Law, etc. We perform ablation studies on selecting the transformer models and how their individual complexity scores are aggregated to get the resulting complexity scores. Our method\footnote{The code is available at \url{https://github.com/abhi1nandy2/CS60075-Team-2-Task-1}} achieves a best Pearson Correlation of $0.784$ in sub-task 1 (single word) and $0.836$ in sub-task 2 (multiple word expressions). 
\end{abstract}


\section{Introduction}

Complex words hinder the readability of a text, as discussed in ~\cite{Dubay2004}. To mitigate this problem, there is a necessity of lexical simplification~\cite{leroy2013user}, and predicting the complexity of words is an integral part of this process. 

Language Models learn the probability of co-occurrence of words in a corpus. They have been used for various sentence completion and text-based classification tasks. The first language models were n-gram Markov Models~\cite{HMM}, which performed well for tasks that did not require very long-range dependencies. Then came RNNs~\cite{rnn}, LSTMs~\cite{lstm} and GRUs~\cite{gru}, which were able to understand longer contexts, but struggled with long paragraphs due to the vanishing gradient problem. Transformers~\cite{transformers} were a task-agnostic solution that performed better due to the presence of Attention Layers between hidden layers of the neural network, which helped the layers of the neural network to look at the entire input at once. Transformers can perform very well on a broad suite of tasks by fine-tuning on a small number of task-specific samples. The intuition behind using such transformer-based language models for Lexical Complexity Prediction (LCP) was - transformer models pre-trained on different corpora would mimic annotators (of the CompLex Dataset~\cite{complex}) having different backgrounds. Since the final score is an aggregation of the annotation scores given by annotators, we aggregate the various scores that are given as outputs by the transformer-based models fine-tuned on the CompLex Dataset.

The rest of the paper is organized as follows. Section~\ref{sol_overview} gives an overview of our solution approach, Section~\ref{data} talks about the corpora used for pre-training and the dataset used for fine-tuning for the LCP task, Section~\ref{exp_res} discusses the experimental settings, baselines used, and a comparison and analysis of the results and Section~\ref{conclusion} gives a conclusion.

\section{Solution Overview}
\label{sol_overview}
\subsection{Model Architecture}

\begin{figure}[t]
    \centering
    \includegraphics[scale=0.35]{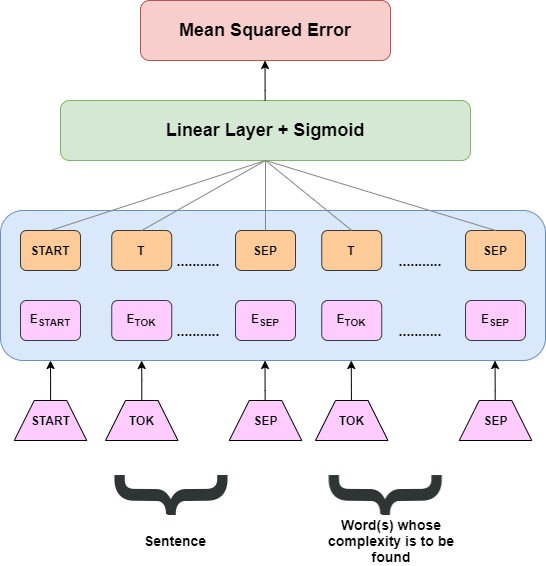}
    \caption{General Block Diagram of the Transformer}
    \label{fig:transformer_reg}
\end{figure}

We use several transformer models. The general block diagram of such a model is shown in Fig.~\ref{fig:transformer_reg}. The input to a model is the tokenized form of a sentence, and the tokenized form of the word/multi-word expression whose complexity score is to be predicted (separated by special tokens), and the target output is the complexity score. Each model consists of a transformer encoder, having the architecture of either BERT~\cite{bert} or RoBERTa~\cite{roberta}, followed by a linear layer and a sigmoid activation layer so that the output is squashed in the range $(0, 1)$. Sigmoid Activation Function is applied, as the target complexity score is a value between $0$ and $1$. To compute loss for backpropagation, the mean squared error loss function is used, as the problem is, as such, a regression problem. 

\subsection{Pre-training the transformer on text corpora}

In order to initialize the weights and the embeddings of the transformer encoder, it is pre-trained on large text corpora so that it has syntactic, lexical and semantic knowledge before fine-tuning on the task-specific data. This is done for two reasons - (1) To increase the rate of convergence towards the lexical complexity prediction task (2) To mimic an annotator from a particular background. 

In order to pre-train a transformer, specific pre-training tasks are performed. If the transformer being used is RoBERTa, Masked Language Modelling (MLM) is performed, where $15\%$ of all the tokens are randomly replaced by a $<MASK>$ token. Such a masked sentence is provided as input to the transformer language model, and a Softmax Layer activation Function is applied for the output corresponding to the masked token to find out the probabilities of various tokens in the vocabulary being in the place of the $<MASK>$ in the original, unmasked sentence. The target is the actual token that was masked. A cross-entropy loss function is used to calculate the loss that is backpropagated. In the case of the BERT Transformer, in addition to the MLM pre-training task, Next Sentence Prediction (NSP) Task is also performed. Two sentences are taken from the corpus, where either one sentence follows the other, or the two sentences are far apart. The output is either $1$ corresponding to the sentences being adjacent to each other, and $0$ being the case when they are far apart. Both the cases have the same number of samples while training. The output corresponding to the $START$ (here, $<CLS>$) token is passed through a  linear layer to get a $2$x$1$ shaped vector, which is then followed by a Softmax Layer, thus giving probabilities of whether the second sentence comes after the first one or not. 
\section{Data}
\label{data}
\subsection{Data used for pre-training}
\label{pre_train_data}
Since we require several transformer language models pre-trained on a wide variety of corpora, we make it a point that we have transformers pre-trained on text corpora from which the CompLex Dataset has been extracted. These corpora are - (1) World English Bible Translation~\cite{christodouloupoulos2015massively} (We used the data found in this link~\footnote{\url{https://www.kaggle.com/oswinrh/bible}}) (2) English part of the Europian Parliament Proceedings from europarl~\cite{koehn2005europarl} (3) CRAFT corpus~\cite{bada2012concept} of bio-medical domain. We pre-train three RoBERTa language models on these three corpora (initialized by weights from~\cite{roberta}) using the MLM pre-training task.

\subsection{Data used for fine-tuning}
For fine-tuning, we do not use any external data other than the datasets that have been provided for both the sub-tasks~\footnote{\url{https://github.com/MMU-TDMLab/CompLex}}.

\section{Experiments and Results}
\label{exp_res}
\begin{table*}[htbp]
\centering
\resizebox{\textwidth}{!}{\begin{tabular}{l|cc|cc}
\hline
\multicolumn{1}{l}{} & \multicolumn{2}{c}{\textbf{Single word}} & \multicolumn{2}{c}{\textbf{MWE}}                                                                                                                                                                                                                                                    \\ \hline
\multicolumn{1}{c|}{\textbf{APPROACH}}                                                                                                      & \textbf{\begin{tabular}[c]{@{}c@{}}PC\end{tabular}}   & \textbf{MSE} & \textbf{\begin{tabular}[c]{@{}c@{}}PC\end{tabular}}  & \textbf{MSE} \\ \hline
\multicolumn{1}{l|}{\begin{tabular}[c]{@{}l@{}}xgb-A\end{tabular}}                                             & 0.718                      & 0.0078                                                    &0.762                    & 0.0103   \\ \hline
 \multicolumn{1}{l|}{\begin{tabular}[c]{@{}l@{}}xgb-B\end{tabular}}             & 0.741                                                          &  0.0073                  & 0.815                                                          & 0.0083       \\ \hline
 \multicolumn{1}{l|}{\begin{tabular}[c]{@{}l@{}}xgb-C\end{tabular}} & 0.744                                                            & 0.0072       & 0.817                                                                     & $0.0082^*$              \\ \hline
 \multicolumn{1}{l|}{\begin{tabular}[c]{@{}l@{}}BERT-BASE-UNCASED\end{tabular}} & 0.765                                            & $0.007^*$        & 0.791                                                                  & 0.009    \\ \hline
 \multicolumn{1}{l|}{\begin{tabular}[c]{@{}l@{}}BIBLE+EUROPARL+BIOMED (AVG.)\end{tabular}}               &0.753&0.0075                 &0.798&0.0096 \\ \hline
   \multicolumn{1}{l|}{\begin{tabular}[c]{@{}l@{}}BIBLE+EUROPARL+BIOMED (MAX.)\end{tabular}}               & 0.751                                                 & 0.0076                  & 0.788  & 0.0092\\ \hline
 \multicolumn{1}{l|}{\begin{tabular}[c]{@{}l@{}}BEST COMBINATION (AVG.)\end{tabular}}               &\textbf{0.784}&\textbf{0.0066}                 &\textbf{0.836}&\textbf{0.0078} \\ \hline
   \multicolumn{1}{l|}{\begin{tabular}[c]{@{}l@{}}BEST COMBINATION (MAX.)\end{tabular}}               & $0.774^*$                  & 0.0071                  & $0.819^*$  & 0.0091\\ \hline
\end{tabular}}
\caption{Comparing the Pearson Correlation (PC) and Mean Squared Error (MSE) of our methods and the baselines (The entries in \textbf{bold} are the best performing according to the respective column's metrics, while the ones with a $^*$ are the next best ones.)} 
\label{results-1}
\end{table*}

\subsection{Transformer Language Models used}

We use the predictions from $9$ transformer-based language models, $4$ of which have a RoBERTa encoder, and the other $5$ have a BERT-based encoder. $2$ models are pre-trained on general domain corpora like Wikipedia and BooksCorpus, $2$ models on biomedical and clinical data, $2$ models on Europarl data, $1$ on Bible, $1$ on Financial data, and $1$ on scientific papers. Also, $6$ of the pre-trained transformer models were publicly available in the HuggingFace Models Catalog~\footnote{\url{https://huggingface.co/models}}, while the other $3$ were pre-trained by us on the three datasets from which CompLex Dataset is extracted, as mentioned in Section~\ref{pre_train_data}.

\subsection{Training, validation and Test Sets}

For each sub-task, the training and the test sets are the same as those provided for the competition. The trial data given for each sub-task is taken to be the validation data.

\subsection{Hyperparameters}

For pre-training, the RoBERTa transformer language model, a batch size of $16$ is used and is trained up to $1$ epoch. The rest of the parameters are the same as in~\cite{roberta}.

When fine-tuning, irrespective of whether the model has a RoBERTa or a BERT Transformer encoder, the input sequence length is set to $256$, with padding or truncation, as is the case. A learning rate of $2\times10^{-5}$ is used with a batch size of $32$, and a Weighted Adam Optimizer is used. The network is fine-tuned for $4$ epochs. The Pearson Correlation on the validation data is calculated for every epoch, and the checkpoint giving the best Pearson Correlation is regarded as the best checkpoint, which would later be used for predicting outputs on the test data. 

\subsection{Methods of Aggregation used}

In order to aggregate the complexity scores of a particular combination of models, we use the following two strategies - sample-wise average and sample-wise maximum across all transformer models. We then do the same across all permutations, see which combination gives the best test results and report it as the final result.

\subsection{Baselines}

We use XGBoost~\cite{xgb} to perform a boosting-based regression model, with an objective of squared error and other default parameters and hyperparameters over a set of features. The different baselines use different feature sets, which are as follows -

\begin{enumerate}
    \item \textbf{xgb-A} - word length (sum of word lengths in the case of MWE), number of syllables and word frequency (from various text sources)~\cite{robyn_speer_2018_1443582} (average of word frequencies in the case of MWE) of the word/expression whose complexity is to be found, and the type of corpus of the sentence (either Bible, Europarl, or Biomedical).
    \item \textbf{xgb-B} - Concatenation of features of xgb-A and the 50 and 100-dimensional GloVe~\cite{GloVe} word vectors of the word/expression whose complexity is to be found. For the expression, the sum of the GloVe Vectors of the individual words would be taken.
    \item \textbf{xgb-C} - Concatenation of features of xgb-B and the probabilities of the word/expression whose complexity is to be found given the sentence with that word/expression that is masked, where the probabilities are predicted by different transformer-based masked language models pre-trained on different corpora.
    
    \textbf{Note: } The probability of the word/token given the masked sentence is approximated as the product of the probabilities of predicting each token, given other tokens of the sentence are masked.
    E.g., Given a sentence $S$ - "I just love mowing the lawn with a lawn mower." Let's say one is required to find out the complexity of the expression - ``lawn mower". First, `lawn' is masked in $S$, and the probability to predict `lawn' using the transformer model $M$ is found, denoted by $P1$.  Similarly, `mower' is masked in $S$, and the probability to predict `mower' using $M$ is found, denoted by $P2$. Resultant feature value = $P1 * P2$
\end{enumerate}

\subsection{Results and Discussion}

Table~\ref{results-1} compares the Pearson Correlations (higher the better) and mean squared errors (lower the better) of our best (according to Pearson Correlation) aggregate results (for both average as well as maximum aggregation), some ablations, and the baselines for both the sub-tasks.

Based on the results, we can infer the following -

\begin{enumerate}
    \item \textbf{xgb-B} performs better than \textbf{xgb-A}, suggesting that, GloVe Word Vector features perform a vital role in complexity prediction, as they contain some contextual information regarding the word.
    \item \textbf{xgb-C} performs the best among the baselines, as it also considers the probabilities of predicting the masked tokens whose complexity is found, adding to the contextual information.
    \item Fine-tuning \textbf{BERT-BASE-UNCASED} transformer model for the LCP task performs better than the best baseline in case of sub-task 1, which could be attributed to the reason that, fine-tuning attention-based transformer models captures even more contextual information than the baselines.
    \item Fine-tuning and aggregating RoBERTa Transformer models pre-trained on the three corpora from which the CompLex Dataset was extracted (\textbf{BIBLE+EUROPARL+BIOMED}), still gives better results than the baselines (except for \textbf{xgb-B} and \textbf{xgb-C} in case of sub-task 2), but performs inferior as compared to \textbf{BERT-BASE-UNCASED} model for single word sub-task, while performing almost similar in case of Multi-Word Expressions sub-task. Also, the average aggregation performs better than the maximum aggregation.
    \item The combination of transformer models that gives the best results upon aggregation (\textbf{BEST COMBINATION}), consists of $3$-$4$ different transformer models fine-tuned on the dataset, suggesting that, \textbf{transformer models pre-trained on domains related as well as unrelated to the dataset (such as Financial Data, Legal Data), are able to best mimic annotators coming from various backgrounds.} Even in this case, average aggregation performs better than maximum aggregation.
    \item If we consider the evaluation metrics of Pearson Correlation (PC) and Mean Squared Error (MSE), it can be seen (especially in the single-word sub-task) that they are negatively correlated, as is expected.
\end{enumerate}

\section{Conclusion}
\label{conclusion}
We show that aggregating the results of various fine-tuned transformer models pre-trained on various corpora from different domains gives high Pearson Correlation and low mean squared errors compared to individual transformers and regression models using attributes such as hand-crafted features, word embeddings, transformer-based language model prediction probabilities, etc. This shows that transformer-based language models, each pre-trained on a different text corpus, can better imitate annotators of the dataset, who come from diverse backgrounds and prior knowledge.

\bibliographystyle{unsrt}


\end{document}